\documentclass[conference]{IEEEtran}
\usepackage{cite}
\usepackage{amsmath,amssymb,amsfonts}
\usepackage{algorithmic}
\usepackage{graphicx}
\usepackage{textcomp}
\usepackage{booktabs}
\usepackage{multirow, makecell}
\usepackage{float}
\usepackage{xcolor}
\def\BibTeX{{\rm B\kern-.05em{\sc i\kern-.025em b}\kern-.08em
    T\kern-.1667em\lower.7ex\hbox{E}\kern-.125emX}}
\begin{document}

\title{Fine-tuning Strategies for Domain Specific Question Answering under Low Annotation Budget Constraints}

\author{\IEEEauthorblockN{1\textsuperscript{st} Kunpeng GUO}
\IEEEauthorblockA{\textit{The QA Company SAS } \\
\&
\textit{Laboratoire Hubert Curien, UMR CNRS 5516}\\
Saint-Etienne, France \\
kunpeng.guo@univ-st-etienne.fr}
\and
\IEEEauthorblockN{2\textsuperscript{nd} Dennis Diefenbach}
\IEEEauthorblockA{\textit{The QA Company SAS} \\
\&
\textit{Laboratoire Hubert Curien, UMR CNRS 5516}\\
Saint-Etienne, France \\
dennis.diefenbach@the-qa-company.com}
\and

\IEEEauthorblockN{3\textsuperscript{rd} Antoine Gourru}
\IEEEauthorblockA{\textit{Laboratoire Hubert Curien, UMR CNRS 5516} \\
Saint-Etienne, France \\
antoine.gourru@univ-st-etienne.fr}
\and
\IEEEauthorblockN{4\textsuperscript{th} Christophe Gravier}
\IEEEauthorblockA{\textit{Laboratoire Hubert Curien, UMR CNRS 5516} \\
Saint-Etienne, France \\
christophe.gravier@univ-st-etienne.fr}
}

\maketitle

\begin{abstract}
The progress introduced by pre-trained language models and their fine-tuning has resulted in significant improvements in most downstream NLP tasks.
The unsupervised training of a language model combined with further target task fine-tuning has become the standard QA fine-tuning procedure. In this work, we demonstrate that this strategy is sub-optimal for fine-tuning QA models, especially under a low QA annotation budget, which is a usual setting in practice due to the extractive QA labeling cost. We draw our conclusions by conducting an exhaustive analysis of the performance of the alternatives of the sequential fine-tuning strategy on different QA datasets. \\
Based on the experiments performed, we observed that the best strategy to fine-tune the QA model in low-budget settings is taking a pre-trained language model (PLM) and then fine-tuning PLM with a dataset composed of the target dataset and SQuAD dataset. With zero extra annotation effort, the best strategy outperforms the standard strategy by 2.28\% to 6.48\%.
Our experiments provide one of the first investigations on how to best fine-tune a QA system under a low budget and are therefore of the utmost practical interest to the QA practitioners.
\end{abstract}

\section{Introduction}\label{sec:introduction}
In the recent few years, transformer-based language models like BERT~\cite{devlin2018bert}, RoBERTa~\cite{liu2019roberta}, T5~\cite{raffel2019exploring}) and GPT-3~\cite{brown2020language}, have played a vital role in the Natural Language Processing (NLP) community. They ever reached a wider audience after OpenAI released an impressive demonstrator, ChatGPT, fine-tuned for conversation and prompted generation. Trained on vast amounts of unsupervised data, these foundation models have become the \textit{de facto} starting point for modern NLP pipelines.

Consequently, we focus on encoder-based foundation models in this work and specifically how to adapt them for NLP downstream tasks. Trained on vast amounts of unsupervised data, these foundation models have become the \textit{de facto} starting point for modern NLP pipelines.
The reason is that the adaptability to new tasks of these so-called \textit{foundation models}~\cite{bommasani2021opportunities} has led to substantial improvements in many NLP downstream tasks, such as sequence classification~\cite{gonzalez2020comparing}, text summarization~\cite{miller2019leveraging}, text generation~\cite{raffel2019exploring} and question answering~\cite{yang2019end}.

However, this adaptability comes at a cost: adapting foundation models to a specific and complex task requires a significant amount of annotated samples in order to fine-tune those models to the task at hand~\cite{antonello-etal-2021-selecting}. 
In practice, the training datasets for domain-specific tasks are usually rather small due to budget constraints. 
Having access to hundreds of labeled samples for a task is common and is not tagged as a few-shots scenario, yet the limited annotation budget still makes the fine-tuning task tedious. 

To circumvent this issue, a double fine-tuning step is usually introduced. 
It consists of fine-tuning the pre-trained foundation model on a large-scale training dataset that is as close as possible (domain and objective) to the target task and is then further fine-tuned on the given domain/task for which training data is scarce. %
The result is a Pre-trained Language Model (\texttt{PLM}) like BERT~\cite{devlin2018bert}, trained on masked language modeling or text generation task, that is then fine-tuned on a more specific large-scale task (\texttt{LM'}), and ultimately refined on the domain/task at hand (\texttt{LM''}). Note that all these steps are applied sequentially. 
%
%
In this work, we explore how to best fine-tune models for domain-specific extractive Question Answering (QA) with limited training data.

In the double fine-tuning step stated above, practitioners usually leverage the Stanford Question Answering Dataset (SQuAD)~\cite{rajpurkar2016SQuAD} which is a high-quality QA dataset that covers diverse knowledge for the \texttt{PLM} to train on.
Nonetheless, in many real-life scenarios, specific-domain QA has a range of field applications that is narrower than SQuAD and may not appear in the SQuAD training data. This calls for building a domain-specific dataset to further fine-tune a QA model for the domain at hand to produce a QA model \texttt{LM''}. This last fine-tuning step is domain-dependent, and the practitioner's goal is also to ultimately keep the number of annotated training samples low - they are under a low annotation budget constraint. It's worth mentioning that, for extractive QA, annotating $200$ examples is already a time-consuming work: the collection of question and answer data requires the annotator to read and understand the text in order to ensure the reasonableness of the marked answers. 
%
%

%

In this paper \texttt{budget} refers to the number of domain-specific annotations available at the time of fine-tuning \texttt{LM''}. 
We therefore look for the best fine-tuning strategy for domain-specific QA under the given budget. 

In this work, we focus on encoder-based foundation models only, as the auto-regressive nature of modern models \cite{touvron2023llama} can still lead to hallucinations, which can be problematic for scenarios that prioritize authoritative answers. Additionally, recent large language models (LLMs) are either unavailable or difficult to fine-tune, and their zero-shot performance is not yet comparable to state-of-the-art encoder-only approaches. \cite{kocoń2023chatgpt}. 

Our contribution is a study of the different strategies one can use to fine-tune a domain-specific extractive QA model. 
This study is exhaustive as we report experiments for $108$ different strategies, applied to $4$ different datasets (we discussed $432$ trained models, each ran $5$ times, see~Section~\ref{sec:experiments}). We provide a complete protocol and evaluation scheme freely available to the community\footnote{code and dataset are available in a GitHub repository, private during the review process}. 
Based on these contributions, we explored different low annotation budget scenarios for which our findings are as follows: 
\begin{itemize}
    \item We demonstrate that the standard sequential QA fine-tuning strategy is sub-optimal for QA under a budget 
    \item Contrary to reasonable expectations, fine-tuning the text encoder using masked language modeling on domain corpus prior to task fine-tuning does not provide an improvement (we even consistently observed a slight degradation of performance)
    \item A very low annotation budget goes a long way, that is $200$ annotated QA pairs is very efficient with respect to the annotation required
    \item We demonstrate that is it better to go either with a small annotation budget with a careful choice of fine-tuning strategies or to go for more than $1,600$ annotations. Anything doubling of the annotation budget in between only results in a $2\%$ improvement in rare cases.
\end{itemize}

\section{Related Work}\label{sec:relatedWork}
\begin{figure}[t]
    \includegraphics[width=\linewidth]{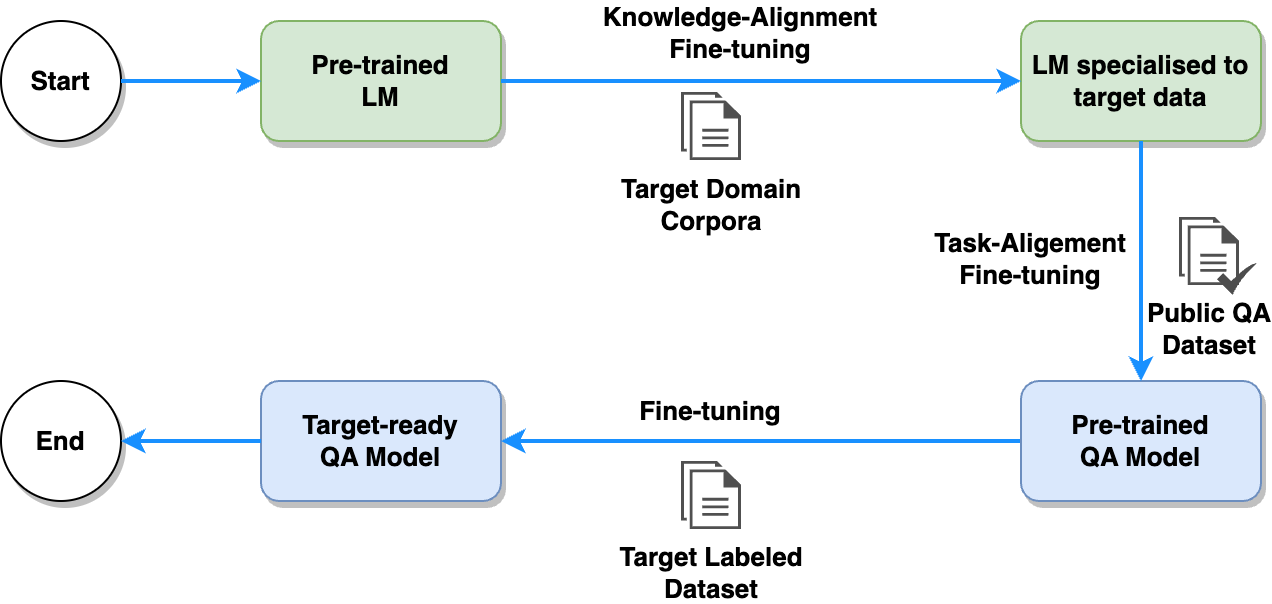}
    \caption{Mainstream methods for QA fine-tuning.}
    \label{fig:standard-finetune-viz}
\end{figure}

In Question Answering, there are mainly three fine-tuning strategies to adapt a language model to a specific domain. 
These strategies are non-exclusive so the standard process to create a domain extractive QA system is to apply them as a sequential pipeline as depicted in Figure~\ref{fig:standard-finetune-viz}. 
In what follows, we describe and discuss the related works to each of these fine-tuning steps.

\subsection{Knowledge-Alignment Fine-tuning}\label{subsec:knowledgeAlignmentFinetune}
Knowledge-Alignment Fine-tuning aims to integrate information about the underlying text corpus into the LM. It is often achieved using the masked language modeling task, inherited from the LM pre-training objective. It helps align the knowledge from the target domain which can be substantially different from what the used LM is pre-trained on. For different NLP tasks, this fine-tuning strategy has shown performance improvements. For example, \cite{Lee_2019} fine-tunes BERT via knowledge-alignment on Biological corpora (PubMed). The corresponding model BioBERT,  can outperform the BERT model in many biomedical text mining tasks like Named Entity Recognition (NER), Relation Extraction (RE), and QA. Similarly, ~\cite{nguyen2020bertweet} generates BERTweet by knowledge-alignment fine-tuning with 850M English tweets. The resulting model gets improvements in part-of-speech tagging, NER, and text classification. 
Nonetheless, it has been shown in~\cite{zhao-bethard-2020-berts} that the benefits vary depending on the task and on the flavor (base or large) version of BERT~\cite{devlin2018bert} and RoBERTa~\cite{liu2019roberta} models. 
\cite{edwards-etal-2020-go} also reports difficulties to fine-tune a BERT model with a limited domain corpus, which is usually the case for domain-specific extractive QA. 

\subsection{Task-Alignment Fine-tuning}\label{subsec:taskAlignment}
Task-Alignment Fine-tuning aims at adapting the pre-trained LM to the target task, that is extractive QA in the scope of this work. 
Generally, publicly available large datasets are used for this purpose. In the QA domain, the dataset of choice is SQuAD~\cite{rajpurkar2016SQuAD} as it contains more than 100k questions on significantly different topics.
In order to obtain a neural extractive QA model, the LM is used as a text encoder, then two independent softmax layers are added on top of it in order to predict the start index and stop index of the answer span~\cite{devlin2018bert}. 
The added layers are shallow, the LM weights are not frozen, and the training, therefore, updates the LM parameters with respect to the extractive QA task. 
This fine-tuning strategy is for example used in ~\cite{Kratzwald,moller-etal-2020-covid}. 
However, in~\cite{merchant-etal-2020-happens}, the authors demonstrate that SQuAD-based fine-tuning involves only shallow changes to the LM and mostly to its top layers. 
In~\cite{cui-etal-2019-fine} the authors try to alleviate this issue by introducing sparse attention in BERT attention heads when fine-tuning -- this however comes to a complexity cost and a modest improvement on SQuAD fine-tuning. 
%
%
Few methods take interest in finding significantly different alternatives, but in~\cite{khashabi2020unifiedqa}, the authors do take an opposite stance. They train a unified QA model, where unified means being able to perform multiple forms of QA (extractive, multiple choice, etc.). As the model is trained to generalize to different QA task formats and still performs well on all domain tasks, it is afterward fine-tuned on each target dataset which ultimately leads to as many QA models.
\subsection{Target Data Fine-tuning}
Target Data Fine-tuning is adapting the LM to the target task using target task-labeled training data. This allows considerable performance improvements, but it is limited by the amount of training data. There are different variations of this strategy. Either the model is directly fine-tuned using the target data (as done in ~\cite{Kratzwald,moller-etal-2020-covid}) or it is trained on a mix between general QA questions and the target ones. In the latter case questions from SQuAD can be used to mix as it is done in ~\cite{Kratzwald2019a}. The authors only explore one way to combine these data, and it is therefore not assumed here that this is the best strategy. 

To summarize, the most common fine-tuning strategy used in literature for domain adaptation in QA is as follows: first, the pre-trained language model (\texttt{PLM}) is optionally further pre-trained in an unsupervised fashion on the domain corpus at hand using the masked language modeling task (\texttt{PLM+}), second, the \texttt{PLM} (or optionally \texttt{PLM+}) is fine-tuned on SQuAD via Task-Alignment Fine-tuning (\texttt{LM'}), and third, the network is fine-tuned again on the domain QA pairs annotations that one may have (\texttt{LM''}). 
%
We are not aware of any study that tried to compare the different fine-tuning strategies and also considered several ways to combine SQuAD and domain-specific corpora when fine-tuning domain-specific extractive QA systems. 

\section{Methodology}
We considered the following options: \\
\textbf{MLM Knowledge-Alignment Fine-tuning:} In our experiments, we used the Masked Language Modeling (MLM) task to distill the knowledge of the corpora into the LM as discussed in Section~\ref{subsec:knowledgeAlignmentFinetune}.
To assess whether knowledge-alignment fine-tuning via MLM helps improve performance under low-budget situations, we conduct the experiments both with and without this procedure for all combinations of fine-tuning strategies presented in Section~\ref{subsec:combinations}.\\
\textbf{SQuAD Fine-tuning}: Following explanations from Section~\ref{subsec:taskAlignment}, we include the possible steps to build a pre-trained QA model based on SQuAD. \\
%
\textbf{Target Data Fine-tuning and Domain Drift Boosting:}
Target data fine-tuning (training on the domain labeled QA samples) usually happens after fine-tuning the text encoder on SQuAD, a high-quality rich dataset for aligning the model to the open domain QA setting.  
However, when it comes to a dataset that is substantially different from SQuAD, both in wording and syntax, this method may become undesirable due to the significant domain drift~\cite{elsahar-galle-2019-annotate}. 
Furthermore, it is known that LMs tend to behave unstable \cite{mou2016transferable} and lean to overfit the dataset. Since we are experimenting on low-budget situations, this effect is amplified and should be avoided. 
In order to solve this problem, we explore the option to merge the SQuAD and Target QA datasets together in order to make the fine-tuning process stable and avoid the catastrophic forgetting usually happening in QA fine-tuning. 
The merged fine-tuning approach can benefit from the original hyper-parameters used in SQuAD fine-tuning and bypass the errors that may occur during extensive hyper-parameters searching. 
Ultimately we would want to have as many target samples as general samples, but accumulating high-quality training datasets of SQuAD's size for every domain is expensive and hardly realistic (Section~\ref{sec:introduction}). 
Inspired by the techniques from classification with imbalance classes, we choose to undersample or oversample the datasets in order to put more emphasis on the domain-specific data. In our case, the options available will be either undersampling SQuAD or oversampling the target dataset.
%
The expectation is that the model does not overfit the target training data as it has also to optimize the general QA training samples that are not included in the domain. 
Nonetheless, the merging of both general and target QA samples is rarely used in the literature, and the ratio on how to best merge the general and target datasets is heavily understudied. 
For this reason, we devised different merging options that we will later compare -- we will show that merging is actually the best strategy and that all of the merging options are not equal. \\\\
To best describe these merging options, we will use the following notations. 
Let $\mathcal{D}_g$ be the general QA training dataset, $\mathcal{D}_t$ the target QA training dataset and $\mathcal{D}_f$ the final merge training dataset we want to build given a dataset merging option. 
We then define the following merging options :
\begin{itemize}
    \item \textbf{TargetQA}, that is only the target samples -- in other words no merge, s.t. $\mathcal{D}_f = \mathcal{D}_t$
    \item \textbf{MP}, Merge Partial SQuAD based on a a 1:1 merge. Since $|\mathcal{D}_g| >> |\mathcal{D}_t|$, we take all samples from $\mathcal{D}_t$, and we sample $n$ samples from $\mathcal{D}_g$, s.t.  
    $\mathcal{D}_f = \mathcal{D}_t \cup \{s_1, \ldots, s_n\}$ where $s_i  \overset{\text{i.i.d.}}{\sim} U(\mathcal{D}_g$) and $U(\mathcal{D}_g)$ denotes the uniform distribution over the set $\mathcal{D}_g$ and $n = |\mathcal{D}_t|$.
    \item \textbf{MPO}, Merge Partial SQuAD with Oversampling is close to the previous \textbf{MP} strategy, but in \textbf{MPO} we sample three times the set $\mathcal{D}_t$ so that the resulting merged QA training dataset $\mathcal{D}_f$ is twice larger -- in \textbf{MPO} the model will see three times more the target samples (as the best-reported value in the work~\cite{Kratzwald2019a}) in order to amplify the learning signals for the target domain while still having to satisfy the samples sampled from $\mathcal{D}_g$. More formally in that merging option: $\mathcal{D}_f = \mathcal{D}_t \cup \mathcal{D}_t \cup \mathcal{D}_t \cup \{s_1, \ldots, s_{n}\}$ where $s_i \overset{\text{i.i.d.}}{\sim} U(\mathcal{D}_g$).
    \item \textbf{MW}, Merge Whole SQuAD, that is the union of both training dataset s.t. $\mathcal{D}_f = \mathcal{D}_t \cup \mathcal{D}_g$. For that merging option, the QA model will be trained on much more training samples for better QA in general, at the expense to learn from a weaker signal coming from the target task. It is interesting to note that under this merging option, the training data are absolutely the same as the mainstream sequential approach to fine-tuning, although there are not drawn sequentially when training but mixed in a single training step. This single difference,  embarrassingly simple, accounts however for $5$ and up to $10$ macro-f1 increase for all datasets but one when the budget is set to $100$ annotations. 
    \item \textbf{MWO}, Merge Whole SQuAD with Oversampling is close to the previous \textbf{MW} strategy but we do the same oversampling as \textbf{MPO}, then we have  $D_{f} = D_{t} \cup D_{t} \cup D_{t} \cup {D_g}$ where $\mathcal{D}_g$ keeps its original size.
\end{itemize}
We also considered a curriculum learning approach~\cite{curriculum}, in which more simple QA pairs will be used and we would introduce more and more difficult QA samples as the training progresses. 
Since evaluating the QA pair difficulty is not trivial, we explore this possibility by brute force as we generated a large number of experiments with different QA pairs splits that are introduced as the training progress. We observed no significant changes in the target model performances, suggesting that either a curriculum approach is not applicable here, or that there is only a very limited number of QA pair sequences that can actually serve a curriculum learning approach. While this was not the primary focus of our work, the existence or nonexistence of such ``golden sequences" has yet to be investigated. 
Moreover, note that we propose merging options in this paper while, as stated in Section~\ref{sec:relatedWork}, sequential transfer learning (PLM → SQuAD → TargetQA) is the go-to method used in most, if not close to all, QA model fine-tuning pipeline in practice. 


\subsection{Fine-tuning combinations}\label{subsec:combinations}

As stated above, there are a series of options that we can choose to improve the performance of QA models fine-tuning. Combining the options in a different manner lead to as many fine-tuning strategies. In our experiments, we include strategies that can reasonably yield fine-tuning improvements -- we especially discard the combination that first performs fine-tuning on the target dataset and then on SQuAD. 
The meaningful strategies of fine-tuning options we consider for extractive QA fine-tuning in this work are the following: 
\begin{itemize}
    \setlength
    \item PLM → SQuAD
    \item PLM → TargetQA
    \item PLM → SQuAD → TargetQA
    \item PLM → SQuAD → MP
    \item PLM → SQuAD → MPO
    \item PLM → MP
    \item PLM → MPO
    \item PLM → MW
    \item PLM → MWO
\end{itemize}
All the methods listed above will experiment with knowledge-alignment fine-tuning (unsupervised masked language modeling on the target document corpus) as well so that we end up with $18$ different fine-tuning strategies.

\subsection{Datasets}\label{subsec:datasets}
\textbf{SQuAD} is a QA dataset introduced in ~\cite{rajpurkar2016SQuAD}. The dataset contains 100,000 triplets (passages, questions, answers). The passages come from 536 Wikipedia articles. The questions and answers are constructed mainly by crowdsourcing: annotators are allowed to ask up to 5 questions on an article and need to mark the correct answers in the corresponding passage. The major difference between SQuAD and previous QA datasets such as CNN/DM~\cite{hermann2015teaching}, CBT~\cite{hill2016goldilocks}, etc, is that the answers in SQuAD are not a single entity or word, but maybe a phrase, which makes its answers more difficult to predict. \\
%
As target domain QA datasets, we consider the following four domain-specific datasets: \\
\textbf{COVID-QA}~\cite{moller-etal-2020-covid} is a question answering dataset on COVID-19 publications. The dataset contains 147 scientific articles. The quality of the dataset is assured as all the question-answer pairs are annotated by 15 experts with at least a master's degree in biomedicine. \\
\textbf{CUAD-QA}~\cite{hendrycks2021cuad} contains questions about legal contracts in the commercial domain. The corpus, curated and maintained by the Atticus Project, contains more than $13,000$ annotations in $510$ contracts. The original task is to highlight important parts of a contract that are necessary for humans to review. We convert it into a question-answering task in SQuAD fashion. The passages to select are lengthy compared to SQuAD paragraphs. \\
\textbf{MOVIE-QA} contains questions about movie plots extracted from Wikipedia. We constructed the dataset from the DuoRC~\cite{saha2018duorc} dataset. The original dataset is an English language dataset of questions and answers collected from crowd-sourced AMT workers on Wikipedia and IMDb movie plots. It contains two sub-datasets SelfRC and ParaphraseRC. We sampled questions from the SelfRC since the answers of the ParaphraseRC sub-set are paraphrased from the plot of the movie. \\
\textbf{KG-QA}\label{subsub:KGQA} is a dataset that we constructed from the Wikidata knowledge base. It contains keyword questions that are constructed semi-automatically as it is done in a knowledge extraction task using QA techniques borrowed from~\cite{Kratzwald}. More specifically, We extracted 982 entities accompanied by their related
Wikipedia pages containing predicates like game platform, developer, game mode and etc.\\
%
Those four datasets were chosen so that they represent different domains and contain question/answer/context with different characteristics. For the purpose of budget analysis, we randomly sampled $2,000$ examples from each dataset for comparison, and we split our datasets in a 5-fold cross-validation manner to reduce randomness in our experiments. All datasets are in the SQuADv1.1 version i.e. all the questions are answerable.
\subsection{Budget Setting}\label{subsec:budgetsize}
Inspired by the training size analysis in ~\cite{edwards-etal-2020-go}, we choose the following experiment budget sizes: $100$, $200$, $400$, $800$, $1200$, $1600$. Those examples are randomly extracted from the training set.
As for the evaluation of the QA systems in different situations, we use the hold-out test sets ($400$ examples) for comparison.
\subsection{Dataset Analyses}\label{sec:analysis}
In the following, we are trying to measure the gap between SQuAD and each dataset from different perspectives.
\subsubsection{Corpus Analysis}\label{sec:corpus_analysis}
\begin{table}[H]
    \caption{Vocabulary overlap (\%) between domain-specific datasets and general dataset SQuAD.}
    \centering
    \large
     \resizebox{1.0\linewidth}{!}{
    \begin{tabular}{lcccc}
        \toprule
         {} & \textbf{COVID-QA} & \textbf{CUAD-QA} & \textbf{MOVIE-QA} & \textbf{KG-QA} \\
        \midrule
        \textbf{SQuAD} & 36.0 & 34.8 & 41.4 & 50.6 \\
        \bottomrule
    \end{tabular}
    }
    \label{tab:vocabulary-overlap-viz}
\end{table}
\noindent
\textit{Domain Similarity.} We compute a domain similarity metric to objectively identify if a dataset is close or far from SQuAD. We consider the top-10K most frequent unigrams (stop-words excluded) in each dataset and compute the vocabulary overlap (see Table~\ref{tab:vocabulary-overlap-viz}). We observed that MOVIE-QA and KG-QA have a stronger similarity with SQuAD dataset than the others. This is reasonable since MOVIE-QA and KG-QA are based on movie plots from Wikipedia and Wikipedia pages of video game entities respectively. COVID-QA and CUAD-QA are relatively far from SQuAD since the two domains are very specialized either in biology or legal terms. \\
\begin{table}[H]
    \caption{Characteristics of QA datasets used in our experiments. RoBERTa-PC is RoBERTa Pre-training Corpora (PC) reported here for comparison. Candidates are answer candidates in the corpus.}
    \centering
    \large
    \resizebox{1.0\linewidth}{!}{
        \begin{tabular}{lcccc}
        \toprule
            \thead{\textbf{Dataset}} & 
            \thead{\textbf{Avg tokens} \\ \textbf{per question}} & 
            \thead{\textbf{Avg tokens} \\ \textbf{per answer}} & 
            \thead{\textbf{Avg tokens} \\ \textbf{per document}} & 
            \thead{\textbf{Corpus size}}
             \\  \midrule
             {\textbf{RoBERTa-PC}} & {-} & {-} & {-} & {160Gb} \\
             \addlinespace
             {\textbf{SQuAD}} & {10.06} & {3.16} & {116.64} & {13Mb} \\
             \addlinespace
             {\textbf{COVID-QA}} & {9.43} & {13.93} & {4021.83} & {50Mb}\\
             \addlinespace
             {\textbf{CUAD-QA}} & {18.53} & {41.87} & {8428.79} & {153Mb} \\
             \addlinespace
             {\textbf{MOVIE-QA}} & {7.35} & {2.5} & {601.81} & {6.8Mb} \\
             \addlinespace
             {\textbf{KG-QA}} & {3.32} & {1.65} & {1373.65} & {17Mb} \\
             \bottomrule
        \end{tabular}
    }
    \label{tab:datasets-specs}
\end{table}
\noindent
\textit{Corpus size}: The size of the text corpus is shown in Table \ref{tab:datasets-specs}. Note that the corpus size is a fraction of the corpus used for PLMs.
\subsubsection{Question/Answer/Passage Analysis}
Different lengths of questions, answers, and passages can lead to different inference difficulties. Therefore, the length distribution (see in Table~\ref{tab:datasets-specs}) can be a very important metric for evaluating the characteristics of four datasets.
First of all, from the answer length, SQuAD together with KG-QA and MOVIE-QA are dominated by short answers. More specifically, the questions in MOVIE-QA are mostly based on character names or dates of specific events. As for KG-QA, the questions are keyword queries and the answers are constructed with the object entities. With this respect, KG-QA and MOVIE-QA can be considered as SQuAD-like but KG-QA is relatively difficult due to the keyword queries. Besides, all three datasets are based on Wikipedia articles with different granularity: SQuAD is built on top of paragraphs, MOVIE-QA is curated from the movie plots and KG-QA is based on the entire article which can be a bit lengthy.
Second, the passages of CUAD-QA are collected from commercial legal contracts in a specific format. It is substantially different than Wikipedia articles in the way of sentence expression and wording choices. Also, there is an important gap between CUAD-QA and other datasets in answer length, which infers it is a more difficult and absolutely not SQuAD-like dataset. %
For COVID-QA, the dataset is built on top of biological papers, which is also lengthy to infer. 
But for the questions and answers, either in the way of asking questions or the type of the answers, COVID-QA is not far from SQuAD. For that matter, COVID-QA is a dataset similar to SQuAD, but relatively more difficult for inference.

Overall one can say that the gap between COVID-QA and MOVIE-QA with SQuAD is smaller than the two other datasets since the questions and answers length as well as the domain are relatively similar. 

\section{Experiment}\label{sec:experiments}
\subsection{Language Model and fine-tuning strategies}
In our experiments, we use RoBERTa~\cite{liu2019roberta} as our starting PLM. 
RoBERTa is built on BERT: it mainly optimizes key hyper-parameters and simplifies the training objective and training mini-batches size. 
RoBERTa achieves better performance than BERT in many benchmarks like GLUE~\cite{wang2019glue}, SQuAD~\cite{rajpurkar2016SQuAD} and RACE~\cite{lai2017race}, which explains this choice over BERT base or large models. To build extractive QA models, we apply two independent softmax layers for predicting the starting and the ending index of the answer span as in~\cite{devlin2018bert}. Parameters of softmax layers and the PLM are updated when fine-tuning. As for implementation details, we use the pre-trained, $12$-layers, $768$-hidden, $12$-heads, $125$M parameters, RoBERTa base model from HuggingFace hub. AdamW~\cite{loshchilov2019decoupled} is used as the optimizer for fine-tuning with a learning rate set to $3e-5$. The results are reported using 5-fold cross-validation.
We explore $18$ different fine-tuning combinations (see Section~\ref{subsec:combinations}) for $6$ different annotation budget sizes (see Section~\ref{subsec:budgetsize}) over $4$ datasets. Moreover, each unique experiment is actually run $5$ times for different data splits to get significant results. We therefore hereby provide results based on $2,160$ evaluation runs (with $1918$ fine-tuned models). All $2,160$ evaluation runs require $62.5$ days of $4$ Titan XP GPUs to complete.


\subsection{Results}\label{subsec:results}
In what follows, we present the main results and analysis we can deduce from our experiments. It is important to note that we provide a summary table with all our results in Table~\ref{tab:expsResults}. We hereby discuss our findings step by step,  sometimes with a subset of budget sizes, and the interested reader can analyze the complete experiments table in supplementary materials that compile all the  $2,160$ evaluation runs.\\

\begin{table}[H]
    \caption{(Upper part) Comparison between baseline method and the best performant fine-tuning strategy under different budget constraints. (Lower part) Comparison between baseline method and the best in average fine-tuning strategy (RoBERTa-Base-MWO) based on average F1-score of all budgets in all the 4 domains.}
    \centering
    \large
     \resizebox{1.0\linewidth}{!}{
    \begin{tabular}{lccccccc}
        \toprule
         {} & \multicolumn{6}{c}{\textbf{Annotation Budget}} \\
         \cmidrule(lr){2-7}
         {} & \textbf{100} & \textbf{200} & \textbf{400} & \textbf{800} & \textbf{1200} & \textbf{1600} & \textbf{Avg} \\
        \midrule
        \textbf{Baseline strategy} & 56.70 & 61.35 & 64.88 & 67.45 & 68.90 & 69.90 & 65.28 \\
        \textbf{Best performant} & 63.63 & 65.85 & 67.78 & 69.83 & 70.75 & 71.40 & 68.21 \\
        \textbf{Difference} & +6.93 & +4.5 & +2.9 & +2.38 & +1.85 & +1.5 & +2.93 \\
        \midrule
        \textbf{Baseline strategy} & 56.70 & 61.35 & 64.88 & 67.45 & 68.90 & 69.90 & 65.28\\
        \textbf{RoBERTa-Base-MWO} & 63.18 & 65.40 & 67.15 & 68.45 & 69.20 & 69.65 & 67.17\\
        \textbf{Difference} & +6.48 & +4.05 & +2.28 & -1.50 & +0.30 & -0.25 & +1.89\\
        \bottomrule
    \end{tabular}
    }
    \label{tab:gap-best-baseline}
\end{table}

\textbf{(1) The standard fine-tuning strategy for QA is sub-optimal with low training budgets}, and although low training budgets are the \textit{de facto} situations in practice (Section~\ref{sec:introduction}). 
Out of $24$ datasets and budget combinations, it only achieves twice the best performance by a small margin (Table ~\ref{tab:expsResults}).
On the contrary, the performance difference between the mainstream method and the best performant fine-tuning strategy identified is up to  $12.5\%$ for the KG-QA dataset and the budget set to $100$.
The gap is particularly high for low budgets ($k=100$) and tends to be smaller for higher budgets ($k>800$) see in the upper part of Table~\ref{tab:gap-best-baseline}. For a very low budget ($k=100$) the average difference is $6.93\%$, which is substantial. We remind here that i) such an improvement comes at no additional cost for QA practitioners, ii) annotations are very costly as each consists in writing down a question and highlighting an answer after having read and understood a paragraph. \\
Furthermore, based on table~\ref{tab:expsResults}, we can infer which method is the best practice for fine-tuning QA models. First, we calculate the average performance of the 4 domains and further calculate the average of all budget settings for each of the listed methods. It turns out that, overall, \textit{RoBERTa-Base-MWO} is the best strategy. As shown in the lower part of Table~\ref{tab:gap-best-baseline}, this strategy is definitely the go-to strategy in lower-budget settings. It steadily outperformed the baseline strategy by $2.28\%$ to $6.48\%$ from budget $k=100$ to budget $k=400$. However, we observed a drift when budget $k$ is bigger than $800$ examples. the best strategy observed slightly under-performed the baseline method from $0.25\%$ to $1.50\%$. 

Based on all our observations from experiments conducted on the 4 domains, first, we demonstrate that the baseline approach commonly used in recent studies is sub-optimal. On the other hand, we also provide NLP practitioners with the best strategy for fine-tuning QA models under low-budget constraints.
\begin{table}[H]
    \caption{Average performance (\%) difference after MLM procedure evaluated over all the budgets and strategies.}
    \centering
    \large
     \resizebox{1.0\linewidth}{!}{
    \begin{tabular}{lcccc}
        \toprule
         {} & \multicolumn{4}{c}{\textbf{Dataset}} \\
         \cmidrule(lr){2-5}
         {} & \textbf{COVID-QA} & \textbf{CUAD-QA} & \textbf{MOVIE-QA} & \textbf{KG-QA} \\
        \midrule
        \textbf{No MLM} & 55.44 & 38.70 & 78.06 & 77.62 \\
        \textbf{With MLM}    & 52.97 & 38.97 & 77,62 & 63.9  \\
        \textbf{Difference}  & -2.47 & +0.27 & -0.44 & -0.95 \\
        \bottomrule
    \end{tabular}
    }
    \label{tab:difference-after-mlm}
\end{table}

\textbf{(2) Knowledge-Alignment Fine-tuning has limited improvements in domain-specific QA under a budget.} For most of the experiments, we cannot observe that knowledge-alignment fine-tuning (MLM) steadily and repeatedly improves the accuracy of the models, we more often report a consistent slight degradation of performance across all fine-tuning combinations (Table~\ref{tab:difference-after-mlm}). Moreover, over the few occurrences where MLM helps, it does only by a small margin (Table ~\ref{tab:expsResults}). 
While knowledge-alignment fine-tuning was reported to be helpful for other NLP tasks, our experiments show that this is not the case for low annotation budget extractive QA. We associate this with the corpora size of the domain datasets that are several orders of magnitude smaller than the corpora used in other works where MLM was identified to be useful. Large text corpora are rather exceptional in domain-specific QA scenarios, we conclude that MLM fine-tuning is generally not advisable.
\begin{table}[H]
    \caption{Performance difference (\%) between Zero-Shot scenarios and Few-Shot scenarios with low budget ($K=100$) and high budget sizes $K=1,600$}
    \centering
    \large
     \resizebox{\linewidth}{!}{
    \begin{tabular}{lcccc}
        \toprule
         {} & \multicolumn{4}{c}{\textbf{Dataset}} \\
         \cmidrule(lr){2-5}
         {} & \textbf{COVID-QA} & \textbf{CUAD-QA} & \textbf{MOVIE-QA} & \textbf{KG-QA} \\
        \midrule
        \textbf{Zero-shot} & 53.8 & 12.2 & 80.2 & 41.9 \\
        \midrule
        \textbf{Low Budget} & 62.3 & 40.0  & 83.6 & 68.6 \\
        \textbf{Difference} & +8.5 & +27.8 & +3.4 & +26.7 \\
        \midrule
        \textbf{High Budget}   & 67.3 & 50.2  & 85.4 & 82.7  \\
        \textbf{Difference} & +13.5 & +38.0 & +5.2 & +40.8 \\
        \bottomrule
    \end{tabular}
    }
    \label{tab:result-3-difference-after-16x-viz}
\end{table}

\begin{figure*}[t]
    \includegraphics[width=0.75\textwidth]{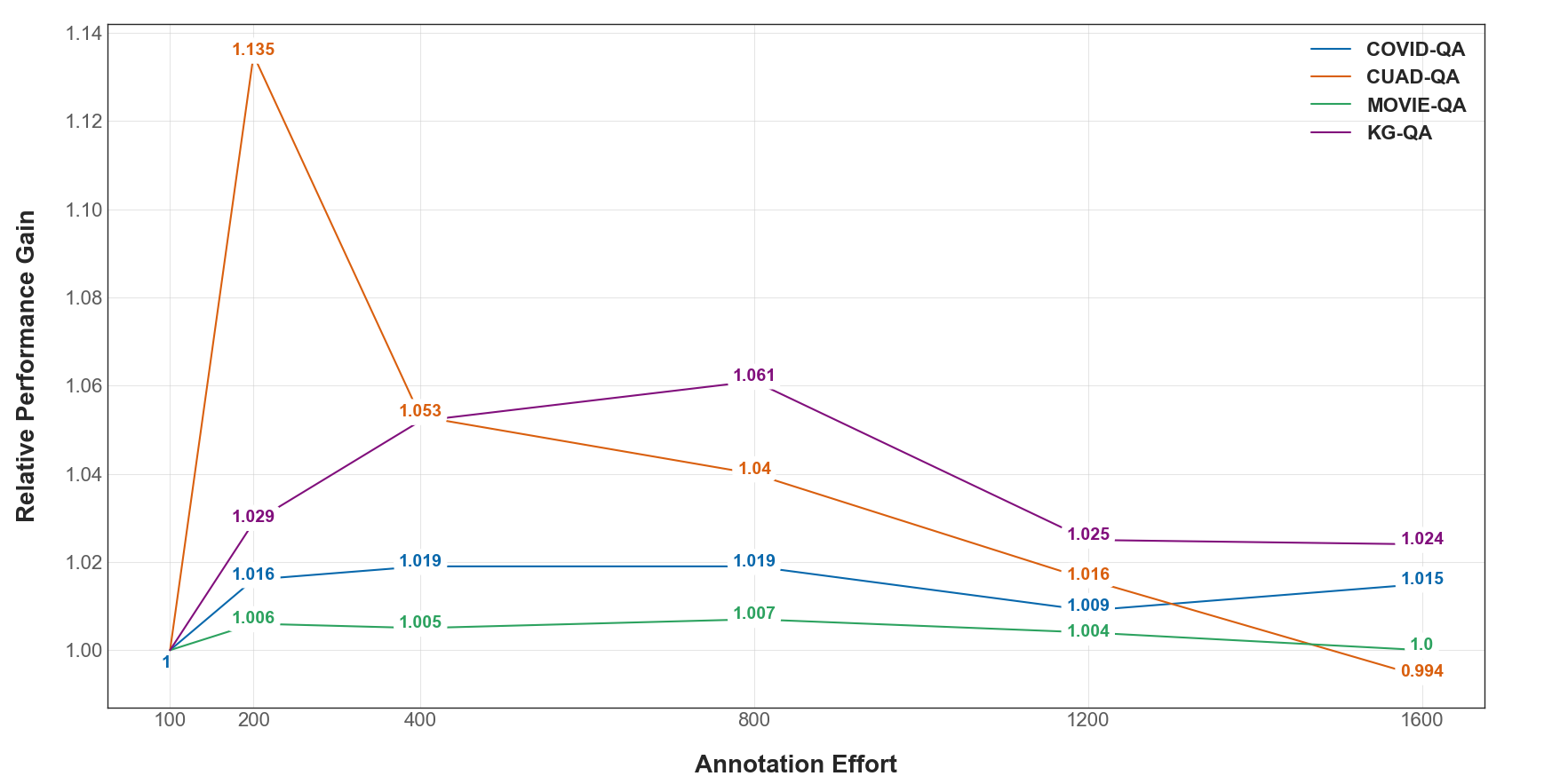}
    \centering
    \caption{Relative performance gain after x16 data collection procedure evaluated over low budget ($K=100$) and high budget sizes ($K=1,600$).}
    \label{fig:result-4-difference-after-16x-viz}
\end{figure*}
%

\textbf{(3) A low annotation budget goes a long way.} Domain-specific training data is assumed to be the best signal to optimize the network in order to achieve better performance. %
We show here that, fortunately, even a small number of samples lead to significant improvements.  To illustrate this, we compare the baseline QA system (RoBERTaBase-SQuAD) with other fine-tuning strategies. Not that we assume that it is possible to devise the best strategies beforehand, which is hardly the case in practice. With this experiment we are quantifying the \emph{expected maximum gain} given all known fine-tuning strategies, assuming access to an oracle providing the best strategy for each fine-tuning strategy. 

In very low-budget scenarios ($k=100$) we observe average performance improvements range between $3.4\%$ and $27.8\%$ (Table~\ref{tab:result-3-difference-after-16x-viz}). For high budgets, it ranges between $5.2\%$ and $40.8\%$. This result is partially consistent with the claim by~\cite{hazen2019towards} that low-budget fine-tuning is actually overestimating the budget in the practical settings since we have just shown that it depends on the domain drift from SQuAD.\\

\textbf{(4) Do not compromise: either go small or big annotation budget.} One of the main issues for the practitioners is to figure out what improvement to expect if one invests more in the budget - we remind here that \texttt{budget} is referred to the number of domain-specific annotations. And building QA annotations are more difficult and therefore costly than text classification for instance. %
We compare the performance improvements that one can achieve by increasing the number of training data -- more training data obviously tend to lead to better models, but we want to measure here how worthy it is to increase the training dataset size. For instance, what are the expectations a practitioner can have if he is willing to double his annotation effort? To answer this question, 
we compare the best performing fine-tuning strategy for each budget (between $K=100$ and $k=1,600$) for the different datasets, assuming that a practitioner is also able to run all strategies to compare them and pick the best one for each budget. We observe the relative gain for each budget jump as reported in Figure~\ref{fig:result-4-difference-after-16x-viz}.

%

From this experiment, we conclude the following. First, providing a small annotation budget ($100$ or $200$) of samples is very efficient with respect to a zero-shot setting (as discussed in the previous experiment). But we also note that doubling the annotation effort leads to only a $1\%$ performance improvement in general and $2\%$ at a maximum. In practice, doubling the amount of extractive QA labels available for target domain fine-tuning is very expensive and therefore does not justify the average $1\%$ improvement (it is also supposed that the experiments were run for all strategies and that the best one was selected, which add to the complexity to benefit fully from these 1 up to 2\% improvement). Complementary, after investing around $10$ times the initial budget, the benefit has accumulated and becomes significant with respect to the effort put into the annotation budget. 

As a rule of thumb, we would advise either opting for a $200$ annotation budget with a careful selection of the \textbf{MWO} fine-tuning strategy or investing for an annotation budget $\geq 1,600$ without the need to explore different fine-tuning strategies in this case. Any effort within the $[200;1,600]$ range implies a weak return with respect to the time and effort to double each time the number of domain annotations. %
%
\section{Conclusion}\label{sec:conclusion}
In this work, we compared different fine-tuning strategies for extractive QA in low-budget scenarios. Our experiments show that the standard fine-tuning strategy for QA is sub-optimal, merge fine-tuning is the most robust and effective fine-tuning strategy, and Knowledge-Alignment Fine-tuning via MLM in small corpora does not yield a significant improvement.

Those are all counter-intuitive results with respect to common practices by the NLP practitioners who usually apply the standard sequential fine-tuning pipeline. We remind the readers that these improvements comes at no overhead cost. Finally, our experiments show what are the performance gains that one can expect by collecting different amounts of training data for different domain-specific QA scenarios depending on similarity with SQuAD.

In future works, we will explore the impact of these different strategies on more recent approaches for fine-tuning Large Language Models, like \cite{hulora}. These new methods allow to achieve good performance while freezing the LLM, that became too big to be fully fine-tuned \cite{touvron2023llama}.

\bibliographystyle{IEEEtran}
\bibliography{ICTAI2023}

\begin{thebibliography}{10}
\providecommand{\url}[1]{#1}
\csname url@samestyle\endcsname
\providecommand{\newblock}{\relax}
\providecommand{\bibinfo}[2]{#2}
\providecommand{\BIBentrySTDinterwordspacing}{\spaceskip=0pt\relax}
\providecommand{\BIBentryALTinterwordstretchfactor}{4}
\providecommand{\BIBentryALTinterwordspacing}{\spaceskip=\fontdimen2\font plus
\BIBentryALTinterwordstretchfactor\fontdimen3\font minus
  \fontdimen4\font\relax}
\providecommand{\BIBforeignlanguage}[2]{{%
\expandafter\ifx\csname l@#1\endcsname\relax
\typeout{** WARNING: IEEEtran.bst: No hyphenation pattern has been}%
\typeout{** loaded for the language `#1'. Using the pattern for}%
\typeout{** the default language instead.}%
\else
\language=\csname l@#1\endcsname
\fi
#2}}
\providecommand{\BIBdecl}{\relax}
\BIBdecl

\bibitem{devlin2018bert}
\BIBentryALTinterwordspacing
J.~Devlin, M.-W. Chang, K.~Lee, and K.~Toutanova, ``{BERT}: Pre-training of
  deep bidirectional transformers for language understanding,'' in
  \emph{Proceedings of the 2019 Conference of the North {A}merican Chapter of
  the Association for Computational Linguistics: Human Language Technologies,
  Volume 1 (Long and Short Papers)}.\hskip 1em plus 0.5em minus 0.4em\relax
  Minneapolis, Minnesota: Association for Computational Linguistics, Jun. 2019,
  pp. 4171--4186. [Online]. Available: \url{https://aclanthology.org/N19-1423}
\BIBentrySTDinterwordspacing

\bibitem{liu2019roberta}
Y.~Liu, M.~Ott, N.~Goyal, J.~Du, M.~Joshi, D.~Chen, O.~Levy, M.~Lewis,
  L.~Zettlemoyer, and V.~Stoyanov, ``Roberta: A robustly optimized bert
  pretraining approach,'' 2019.

\bibitem{raffel2019exploring}
C.~Raffel, N.~Shazeer, A.~Roberts, K.~Lee, S.~Narang, M.~Matena, Y.~Zhou,
  W.~Li, and P.~J. Liu, ``Exploring the limits of transfer learning with a
  unified text-to-text transformer,'' \emph{arXiv preprint arXiv:1910.10683},
  2019.

\bibitem{brown2020language}
T.~B. Brown, B.~Mann, N.~Ryder, M.~Subbiah, J.~Kaplan, P.~Dhariwal,
  A.~Neelakantan, P.~Shyam, G.~Sastry, A.~Askell \emph{et~al.}, ``Language
  models are few-shot learners,'' \emph{arXiv preprint arXiv:2005.14165}, 2020.

\bibitem{bommasani2021opportunities}
R.~Bommasani, D.~A. Hudson, E.~Adeli, R.~Altman, S.~Arora, S.~von Arx, M.~S.
  Bernstein, J.~Bohg, A.~Bosselut, E.~Brunskill \emph{et~al.}, ``On the
  opportunities and risks of foundation models,'' \emph{arXiv preprint
  arXiv:2108.07258}, 2021.

\bibitem{gonzalez2020comparing}
S.~Gonz{\'a}lez-Carvajal and E.~C. Garrido-Merch{\'a}n, ``Comparing bert
  against traditional machine learning text classification,'' \emph{arXiv
  preprint arXiv:2005.13012}, 2020.

\bibitem{miller2019leveraging}
D.~Miller, ``Leveraging bert for extractive text summarization on lectures,''
  \emph{arXiv preprint arXiv:1906.04165}, 2019.

\bibitem{yang2019end}
W.~Yang, Y.~Xie, A.~Lin, X.~Li, L.~Tan, K.~Xiong, M.~Li, and J.~Lin,
  ``End-to-end open-domain question answering with bertserini,'' \emph{arXiv
  preprint arXiv:1902.01718}, 2019.

\bibitem{antonello-etal-2021-selecting}
\BIBentryALTinterwordspacing
R.~Antonello, N.~Beckage, J.~Turek, and A.~Huth, ``Selecting informative
  contexts improves language model fine-tuning,'' in \emph{Proceedings of the
  59th Annual Meeting of the Association for Computational Linguistics and the
  11th International Joint Conference on Natural Language Processing (Volume 1:
  Long Papers)}.\hskip 1em plus 0.5em minus 0.4em\relax Online: Association for
  Computational Linguistics, Aug. 2021, pp. 1072--1085. [Online]. Available:
  \url{https://aclanthology.org/2021.acl-long.87}
\BIBentrySTDinterwordspacing

\bibitem{rajpurkar2016SQuAD}
P.~Rajpurkar, J.~Zhang, K.~Lopyrev, and P.~Liang, ``Squad: 100,000+ questions
  for machine comprehension of text,'' \emph{arXiv preprint arXiv:1606.05250},
  2016.

\bibitem{touvron2023llama}
H.~Touvron, T.~Lavril, G.~Izacard, X.~Martinet, M.-A. Lachaux, T.~Lacroix,
  B.~Rozi{\`e}re, N.~Goyal, E.~Hambro, F.~Azhar \emph{et~al.}, ``Llama: Open
  and efficient foundation language models,'' \emph{arXiv preprint
  arXiv:2302.13971}, 2023.

\bibitem{kocoń2023chatgpt}
J.~Kocoń, I.~Cichecki, O.~Kaszyca, M.~Kochanek, D.~Szydło, J.~Baran,
  J.~Bielaniewicz, M.~Gruza, A.~Janz, K.~Kanclerz, A.~Kocoń, B.~Koptyra,
  W.~Mieleszczenko-Kowszewicz, P.~Miłkowski, M.~Oleksy, M.~Piasecki, Łukasz
  Radliński, K.~Wojtasik, S.~Woźniak, and P.~Kazienko, ``Chatgpt: Jack of all
  trades, master of none,'' 2023.

\bibitem{Lee_2019}
\BIBentryALTinterwordspacing
J.~Lee, W.~Yoon, S.~Kim, D.~Kim, S.~Kim, C.~H. So, and J.~Kang, ``Biobert: a
  pre-trained biomedical language representation model for biomedical text
  mining,'' \emph{Bioinformatics}, Sep 2019. [Online]. Available:
  \url{http://dx.doi.org/10.1093/bioinformatics/btz682}
\BIBentrySTDinterwordspacing

\bibitem{nguyen2020bertweet}
D.~Q. Nguyen, T.~Vu, and A.~T. Nguyen, ``Bertweet: A pre-trained language model
  for english tweets,'' 2020.

\bibitem{zhao-bethard-2020-berts}
\BIBentryALTinterwordspacing
Y.~Zhao and S.~Bethard, ``How does {BERT}{'}s attention change when you
  fine-tune? an analysis methodology and a case study in negation scope,'' in
  \emph{Proceedings of the 58th Annual Meeting of the Association for
  Computational Linguistics}.\hskip 1em plus 0.5em minus 0.4em\relax Online:
  Association for Computational Linguistics, Jul. 2020, pp. 4729--4747.
  [Online]. Available: \url{https://aclanthology.org/2020.acl-main.429}
\BIBentrySTDinterwordspacing

\bibitem{edwards-etal-2020-go}
\BIBentryALTinterwordspacing
A.~Edwards, J.~Camacho-Collados, H.~De~Ribaupierre, and A.~Preece, ``Go simple
  and pre-train on domain-specific corpora: On the role of training data for
  text classification,'' in \emph{Proceedings of the 28th International
  Conference on Computational Linguistics}.\hskip 1em plus 0.5em minus
  0.4em\relax Barcelona, Spain (Online): International Committee on
  Computational Linguistics, Dec. 2020, pp. 5522--5529. [Online]. Available:
  \url{https://aclanthology.org/2020.coling-main.481}
\BIBentrySTDinterwordspacing

\bibitem{Kratzwald}
\BIBentryALTinterwordspacing
B.~Kratzwald, G.~Kunpeng, S.~Feuerriegel, and D.~Diefenbach, ``Intkb: {A}
  verifiable interactive framework for knowledge base completion,'' in
  \emph{Proceedings of the 28th International Conference on Computational
  Linguistics, {COLING} 2020, Barcelona, Spain (Online), December 8-13, 2020},
  D.~Scott, N.~Bel, and C.~Zong, Eds.\hskip 1em plus 0.5em minus 0.4em\relax
  International Committee on Computational Linguistics, 2020, pp. 5591--5603.
  [Online]. Available: \url{https://doi.org/10.18653/v1/2020.coling-main.490}
\BIBentrySTDinterwordspacing

\bibitem{moller-etal-2020-covid}
\BIBentryALTinterwordspacing
T.~M{\"o}ller, A.~Reina, R.~Jayakumar, and M.~Pietsch, ``{COVID-QA}: A question
  answering dataset for {COVID}-19,'' in \emph{Proceedings of the 1st Workshop
  on {NLP} for {COVID-19} at {ACL} 2020}.\hskip 1em plus 0.5em minus
  0.4em\relax Online: Association for Computational Linguistics, Jul. 2020.
  [Online]. Available: \url{https://aclanthology.org/2020.nlpcovid19-acl.18}
\BIBentrySTDinterwordspacing

\bibitem{merchant-etal-2020-happens}
\BIBentryALTinterwordspacing
A.~Merchant, E.~Rahimtoroghi, E.~Pavlick, and I.~Tenney, ``What happens to
  {BERT} embeddings during fine-tuning?'' in \emph{Proceedings of the Third
  BlackboxNLP Workshop on Analyzing and Interpreting Neural Networks for
  NLP}.\hskip 1em plus 0.5em minus 0.4em\relax Online: Association for
  Computational Linguistics, Nov. 2020, pp. 33--44. [Online]. Available:
  \url{https://aclanthology.org/2020.blackboxnlp-1.4}
\BIBentrySTDinterwordspacing

\bibitem{cui-etal-2019-fine}
\BIBentryALTinterwordspacing
B.~Cui, Y.~Li, M.~Chen, and Z.~Zhang, ``Fine-tune {BERT} with sparse
  self-attention mechanism,'' in \emph{Proceedings of the 2019 Conference on
  Empirical Methods in Natural Language Processing and the 9th International
  Joint Conference on Natural Language Processing (EMNLP-IJCNLP)}.\hskip 1em
  plus 0.5em minus 0.4em\relax Hong Kong, China: Association for Computational
  Linguistics, Nov. 2019, pp. 3548--3553. [Online]. Available:
  \url{https://www.aclweb.org/anthology/D19-1361}
\BIBentrySTDinterwordspacing

\bibitem{khashabi2020unifiedqa}
D.~Khashabi, S.~Min, T.~Khot, A.~Sabharwal, O.~Tafjord, P.~Clark, and
  H.~Hajishirzi, ``Unifiedqa: Crossing format boundaries with a single qa
  system,'' 2020.

\bibitem{Kratzwald2019a}
\BIBentryALTinterwordspacing
B.~Kratzwald and S.~Feuerriegel, ``Putting question-answering systems into
  practice: Transfer learning for efficient domain customization,'' \emph{ACM
  Transactions on Management Information Systems}, vol.~9, no.~4, pp.
  15:1--15:20, 2019. [Online]. Available:
  \url{http://doi.acm.org/10.1145/3309706}
\BIBentrySTDinterwordspacing

\bibitem{elsahar-galle-2019-annotate}
\BIBentryALTinterwordspacing
H.~Elsahar and M.~Gall{\'e}, ``To annotate or not? predicting performance drop
  under domain shift,'' in \emph{Proceedings of the 2019 Conference on
  Empirical Methods in Natural Language Processing and the 9th International
  Joint Conference on Natural Language Processing (EMNLP-IJCNLP)}.\hskip 1em
  plus 0.5em minus 0.4em\relax Hong Kong, China: Association for Computational
  Linguistics, Nov. 2019, pp. 2163--2173. [Online]. Available:
  \url{https://aclanthology.org/D19-1222}
\BIBentrySTDinterwordspacing

\bibitem{mou2016transferable}
L.~Mou, Z.~Meng, R.~Yan, G.~Li, Y.~Xu, L.~Zhang, and Z.~Jin, ``How transferable
  are neural networks in nlp applications?'' 2016.

\bibitem{curriculum}
\BIBentryALTinterwordspacing
Y.~Bengio, J.~Louradour, R.~Collobert, and J.~Weston, ``Curriculum learning,''
  in \emph{Proceedings of the 26th Annual International Conference on Machine
  Learning}, ser. ICML '09.\hskip 1em plus 0.5em minus 0.4em\relax New York,
  NY, USA: Association for Computing Machinery, 2009, p. 41–48. [Online].
  Available: \url{https://doi.org/10.1145/1553374.1553380}
\BIBentrySTDinterwordspacing

\bibitem{hermann2015teaching}
K.~M. Hermann, T.~Kočiský, E.~Grefenstette, L.~Espeholt, W.~Kay, M.~Suleyman,
  and P.~Blunsom, ``Teaching machines to read and comprehend,'' 2015.

\bibitem{hill2016goldilocks}
F.~Hill, A.~Bordes, S.~Chopra, and J.~Weston, ``The goldilocks principle:
  Reading children's books with explicit memory representations,'' 2016.

\bibitem{hendrycks2021cuad}
D.~Hendrycks, C.~Burns, A.~Chen, and S.~Ball, ``Cuad: An expert-annotated nlp
  dataset for legal contract review,'' 2021.

\bibitem{saha2018duorc}
A.~Saha, R.~Aralikatte, M.~M. Khapra, and K.~Sankaranarayanan, ``Duorc: Towards
  complex language understanding with paraphrased reading comprehension,''
  2018.

\bibitem{wang2019glue}
A.~Wang, A.~Singh, J.~Michael, F.~Hill, O.~Levy, and S.~R. Bowman, ``Glue: A
  multi-task benchmark and analysis platform for natural language
  understanding,'' 2019.

\bibitem{lai2017race}
G.~Lai, Q.~Xie, H.~Liu, Y.~Yang, and E.~Hovy, ``Race: Large-scale reading
  comprehension dataset from examinations,'' 2017.

\bibitem{loshchilov2019decoupled}
I.~Loshchilov and F.~Hutter, ``Decoupled weight decay regularization,'' 2019.

\bibitem{hazen2019towards}
T.~J. Hazen, S.~Dhuliawala, and D.~Boies, ``Towards domain adaptation from
  limited data for question answering using deep neural networks,'' \emph{arXiv
  preprint arXiv:1911.02655}, 2019.

\bibitem{hulora}
E.~J. Hu, P.~Wallis, Z.~Allen-Zhu, Y.~Li, S.~Wang, L.~Wang, W.~Chen
  \emph{et~al.}, ``Lora: Low-rank adaptation of large language models,'' in
  \emph{International Conference on Learning Representations}.

\end{thebibliography}

\begin{table*}[htbp]
    \caption{Experiment results. $K$ is the budget size. \textit{RoBERTaBase-SQuAD-TargetQA} is the standard sequential fine-tuning method, its results are \underline{underlined} for reference. RoBERTaBase-SQuAD, often referred as the "baseline method" in many benchmarks, reflects how well a SQuAD model generalizes on other QA tasks. Best result for each budget size is given in \textbf{bold}.} 
    \centering
    \resizebox*{!}{0.95\textheight}{
        \begin{tabular}{@{\extracolsep{1pt}}llcccccc}
        \toprule
            {} & {} & \multicolumn{6}{c}{\textbf{MACRO-F1}}\\ 
            \cmidrule(lr){3-8}
            {\textbf{Dataset}} &
            {\textbf{Fine-tune Strategy}} & K = 100 & K = 200 & K = 400 &  K = 800 & K = 1200 & K = 1600 \\
            \midrule
            \multirow{15}{*}{\textbf{COVID-QA}}
            & RoBERTaBase-SQuAD              & 53.8 & 53.8 & 53.8 & 53.8 & 53.8 & 53.8 \\
            & RoBERTaBase-MLM-SQuAD          & 52.9 & 52.9 & 52.9 & 52.9 & 52.9 & 52.9 \\ 
            & RoBERTaBase-TargetQA           & 6.5  & 35.8 & 46.6 & 54.3 & 55.6 & 59.2 \\
            & RoBERTaBase-MLM-TargetQA       & 5.6  & 17.7 & 35.1 & 46.8 & 53.0 & 54.5 \\
            & \textit{RoBERTaBase-SQuAD-TargetQA}     & \underline{55.8} & \underline{58.3} & \underline{61.0} & \underline{63.1} & \underline{64.3} & \underline{64.1} \\
            & RoBERTaBase-MLM-SQuAD-TargetQA & 55.0 & 59.4 & 60.3 & 64.2 & 64.9 & 64.7 \\
            & RoBERTaBase-MP                 & 8.9  & 44.7 & 51.7 & 57.3 & 59.1 & 59.9 \\
            & RoBERTaBase-MLM-MP             & 13.9 & 34.1 & 45.9 & 52.6 & 55.1 & 59.2 \\
            & RoBERTaBase-MPO                & 27.7 & 39.4 & 45.1 & 50.5 & 54.3 & 54.2 \\
            & RoBERTaBase-MLM-MPO            & 18.8 & 29.3 & 37.1 & 47.4 & 51.2 & 52.4 \\
            & RoBERTaBase-SQuAD-MP           & 57.1 & 59.9 & 62.1 & 64.0 & 64.5 & 64.1 \\
            & RoBERTaBase-MLM-SQuAD-MP       & 56.2 & 58.0 & 60.7 & 63.0 & 62.6 & 63.4 \\
            & RoBERTaBase-SQuAD-MPO          & 54.6 & 58.5 & 58.0 & 60.5 & 60.4 & 60.8 \\
            & RoBERTaBase-MLM-SQuAD-MPO      & 53.9 & 56.0 & 57.0 & 59.3 & 58.2 & 60.2 \\
            & RoBERTaBase-MW                 & 60.8 & 62.6 & 64.2 & \textbf{65.7} & \textbf{66.3} & \textbf{67.3} \\
            & RoBERTaBase-MLM-MW             & 60.1 & 63.0 & 63.2 & 64.6 & 65.5 & 65.9 \\
            & RoBERTaBase-MWO                & \textbf{62.3} & \textbf{63.3} & \textbf{64.5} & 64.1 & 63.6 & 64.5 \\
            & RoBERTaBase-MLM-MWO            & \textbf{62.3} & 61.2 & 62.3 & 62.1 & 62.8 & 63.2 \\
            \midrule
            \multirow{15}{*}{\textbf{CUAD-QA}}
            & RoBERTaBase-SQuAD              & 12.2 & 12.2 & 12.2 & 12.2 & 12.2 & 12.2 \\
            & RoBERTaBase-MLM-SQuAD          & 14.2 & 14.2 & 14.2 & 14.2 & 14.2 & 14.2 \\ 
            & RoBERTaBase-TargetQA           & 12.2 & 13.0 & 39.0 & 45.9 & 47.0 & 48.5 \\
            & RoBERTaBase-MLM-TargetQA       & 12.7 & 26.2 & 36.9 & 44.3 & 48.0 & 48.3 \\
            & \textit{RoBERTaBase-SQuAD-TargetQA}     & \underline{35.6} & \underline{42.8} & \underline{45.9} & \underline{47.1} & \underline{48.6} & \underline{\textbf{50.2}} \\
            & RoBERTaBase-MLM-SQuAD-TargetQA & 39.3 & 44.0 & \textbf{47.8} & 48.2 & 48.9 & 49.0 \\
            & RoBERTaBase-MP                 & 12.4 & 34.6 & 43.4 & 45.6 & 48.2 & 49.5 \\
            & RoBERTaBase-MLM-MP             & 18.2 & 31.5 & 40.1 & 46.5 & 47.5 & 49.1 \\
            & RoBERTaBase-MPO                & 21.4 & 35.4 & 40.5 & 44.8 & 45.2 & 45.5 \\
            & RoBERTaBase-MLM-MPO            & 20.0 & 30.5 & 35.0 & 44.2 & 45.0 & 45.5 \\
            & RoBERTaBase-SQuAD-MP           & 38.3 & 42.8 & 46.1 & 49.1 & 49.1 & 48.7 \\
            & RoBERTaBase-MLM-SQuAD-MP       & 38.0 & \textbf{45.4} & 47.2 & \textbf{49.7} & 49.7 & 49.8 \\
            & RoBERTaBase-SQuAD-MPO          & 35.6 & 40.6 & 43.4 & 45.0 & 45.6 & 45.7 \\
            & RoBERTaBase-MLM-SQuAD-MPO      & 35.8 & 41.6 & 44.4 & 45.4 & 46.1 & 46.8 \\
            & RoBERTaBase-MW                 & 35.0 & 42.0 & 45.6 & 48.4 & \textbf{50.5} & 50.0 \\
            & RoBERTaBase-MLM-MW             & 34.5 & 41.5 & 44.6 & 48.7 & 49.4 & \textbf{50.2} \\
            & RoBERTaBase-MWO                & \textbf{40.0} & 45.0 & 46.0 & 46.6 & 47.8 & 47.6 \\
            & RoBERTaBase-MLM-MWO            & 39.1 & 42.7 & 43.3 & 45.8 & 46.2 & 46.6 \\
            \midrule
            \multirow{15}{*}{\textbf{MOVIE-QA}}
            & RoBERTaBase-SQuAD              & 80.2 & 80.2 & 80.2 & 80.2 & 80.2 & 80.2 \\
            & RoBERTaBase-MLM-SQuAD          & 80.0 & 80.0 & 80.0 & 80.0 & 80.0 & 80.0 \\ 
            & RoBERTaBase-TargetQA           & 25.0 & 51.8 & 67.5 & 75.0 & 78.5 & 80.1 \\
            & RoBERTaBase-MLM-TargetQA.      & 25.9 & 44.5 & 54.6 & 74.9 & 77.7 & 79.6 \\
            & \textit{RoBERTaBase-SQuAD-TargetQA}     & \underline{79.3} & \underline{79.9} & \underline{81.9} & \underline{83.2} & \underline{83.4} & \underline{84.0} \\ 
            & RoBERTaBase-MLM-SQuAD-TargetQA & 79.7 & 79.9 & 82.0 & 83.5 & 83.8 & 83.9 \\
            & RoBERTaBase-MP                 & 54.6 & 61.4 & 73.3 & 79.5 & 80.5 & 81.8 \\
            & RoBERTaBase-MLM-MP             & 52.4 & 63.8 & 73.2 & 78.7 & 79.7 & 81.8 \\
            & RoBERTaBase-MPO                & 57.7 & 66.6 & 74.2 & 77.9 & 80.0 & 80.7 \\ 
            & RoBERTaBase-MLM-MPO            & 58.9 & 67.9 & 73.3 & 77.7 & 79.8 & 80.2 \\
            & RoBERTaBase-SQuAD-MP           & 79.2 & 80.4 & 82.2 & 83.5 & 83.6 & 84.6 \\
            & RoBERTaBase-MLM-SQuAD-MP       & 78.5 & 80.9 & 81.0 & 83.0 & 84.0 & 83.3 \\
            & RoBERTaBase-SQuAD-MPO          & 79.7 & 81.3 & 82.4 & 83.3 & 83.2 & 83.4 \\
            & RoBERTaBase-MLM-SQuAD-MPO      & 79.4 & 80.5 & 81.7 & 83.6 & 83.4 & 82.9 \\
            & RoBERTaBase-MW                 & \textbf{83.6} & 83.1 & \textbf{84.5} & 84.4 & 85.1 & 85.0 \\
            & RoBERTaBase-MLM-MW             & 83.0 & 82.9 & \textbf{84.5} & \textbf{85.1} & \textbf{85.4} & \textbf{85.4} \\
            & RoBERTaBase-MWO                & 82.7 & 84.0 & 83.9 & 84.3 & 84.8 & 84.0 \\
            & RoBERTaBase-MLM-MWO            & 83.1 & \textbf{84.1} & 84.3 & 84.9 & 84.5 & 84.5 \\
            \midrule
            \multirow{15}{*}{\textbf{KG-QA}}
            & RoBERTaBase-SQuAD              & 41.9 & 41.9 & 41.9 & 41.9 & 41.9 & 41.9 \\
            & RoBERTaBase-MLM-SQuAD          & 35.9 & 35.9 & 35.9 & 35.9 & 35.9 & 35.9 \\ 
            & RoBERTaBase-TargetQA           & 20.1 & 26.2 & 30.4 & 70.2 & 76.1 & 78.6 \\
            & RoBERTaBase-MLM-TargetQA       & 24.3 & 27.2 & 33.6 & 53.1 & 73.4 & 79.1 \\
            & \textit{RoBERTaBase-SQuAD-TargetQA}     & \underline{56.1} & \underline{64.4} & \underline{70.7} & \underline{76.4} & \underline{79.3} & \underline{81.3} \\
            & RoBERTaBase-MLM-SQuAD-TargetQA & 61.2 & 66.6 & 72.6 & 77.0 & 79.6 & 81.5 \\
            & RoBERTaBase-MP                 & 24.5 & 27.0 & 64.5 & 76.0 & 77.9 & 78.8 \\
            & RoBERTaBase-MLM-MP             & 24.3 & 28.2 & 43.8 & 75.2 & 78.2 & 80.5 \\
            & RoBERTaBase-MPO                & 28.9 & 52.2 & 71.4 & 76.2 & 79.9 & 82.2 \\
            & RoBERTaBase-MLM-MPO            & 40.2 & 40.2 & 70.4 & 77.8 & 79.9 & 82.5 \\
            & RoBERTaBase-SQuAD-MP           & 64.6 & 65.1 & 73.3 & 77.1 & 78.7 & 81.1 \\
            & RoBERTaBase-MLM-SQuAD-MP       & 65.0 & 66.7 & 73.5 & 77.4 & 79.0 & 80.5 \\
            & RoBERTaBase-SQuAD-MPO          & 63.9 & 69.1 & 73.5 & 78.4 & \textbf{80.8} & 82.1 \\
            & RoBERTaBase-MLM-SQuAD-MPO      & 63.9 & 67.8 & 73.5 & 77.5 & 79.8 & 81.7 \\
            & RoBERTaBase-MW                 & 66.1 & 68.2 & 72.3 & 75.8 & 77.7 & 80.4 \\
            & RoBERTaBase-MLM-MW             & 66.2 & 69.2 & 73.5 & 75.8 & 77.8 & 81.0 \\
            & RoBERTaBase-MWO                & 67.7 & 69.3 & 74.2 & \textbf{78.8} & 80.6 & 82.5 \\
            & RoBERTaBase-MLM-MWO            & \textbf{68.6} & \textbf{70.6} & \textbf{74.3} & 78.0 & \textbf{80.8} & \textbf{82.7} \\
            \bottomrule
        \end{tabular}
    }
    \label{tab:expsResults}
\end{table*}

\end{document}